\documentclass[runningheads]{llncs}

\usepackage{eccv}

\usepackage{eccvabbrv}

\usepackage{graphicx}
\usepackage{booktabs}
\usepackage{multirow}
\usepackage{adjustbox}
\usepackage{cuted}
\usepackage{capt-of} 
\usepackage{wrapfig}
\usepackage{graphicx}
\usepackage{amsmath}
\usepackage{algorithm}
\usepackage{algpseudocode}

\usepackage[accsupp]{axessibility}  % Improves PDF readability for those with disabilities.

\usepackage{hyperref}

\usepackage{orcidlink}

\begin{document}

\title{ReflectVLN: Training Vision-Language Navigation Agents with Reflective Reasoning} 

\titlerunning{ReflectVLN}

\renewcommand{\thefootnote}{\fnsymbol{footnote}}

\author{
Jiahang Wang\inst{1}$^{*}$ \and
Yirong Yang\inst{2}$^{*}$ \and
Yanqing Zhu\inst{1}$^{\dagger}$ \and
Minghua Luo\inst{1} \and
Shichao Xie\inst{1} \and
Fei Liu\inst{1} \and
Mu Xu\inst{1}
}

\authorrunning{J.~Wang et al.}

\institute{
Amap, Alibaba Group, China \and
Beihang University
}

\maketitle

\begingroup
\renewcommand{\thefootnote}{}
\footnotetext{$^{*}$ Equal contribution.}
\footnotetext{$^{\dagger}$ Project lead.}
\endgroup

%% Main Paper

% \begin{strip}
% % \vspace{-8.5em}
% \centering
% \includegraphics[width=.95\textwidth]{figures/dummy_pi_0_5.png}
% \captionof{figure}{
% We propose a memory-augmented framework for indoor embodied 3D perception that couples short-term selective read–write aggregation with long-term scene-prior retrieval, enabling robots to robustly accumulate reliable voxel evidence over long-horizon multi-step observations and quickly adapt to new scenes.
% }
% \label{fig1}
% \end{strip}

\begin{abstract}
Existing vision-language navigation methods often couple a VLM with waypoint decoders to produce multi-step action plans, but they typically lack an explicit closed-loop mechanism for tracking semantic progress, diagnosing execution failures, and recovering from error accumulation in long-horizon navigation. To address this gap, we propose ReflectVLN, an agentic VLN framework that organizes decision-making through bidirectionally interactive intention and execution agents. The intention agent performs subtask decomposition and reflection, generating executable subtask descriptions as corrective plans. Conditioned on these descriptions, the execution agent grounds them into short-horizon actions under current observations while monitoring sub-goal progress and detecting off-track behavior. Crucially, ReflectVLN enables closed-loop bidirectional communication: the execution agent emits progress and deviation signals to trigger reflection and subtask updates on demand, and the intention agent returns structured guidance that reconditions subsequent actions for recovery. To encourage temporally coherent decisions with interpretable intermediate rationales, we introduce Action Chain-of-Thought (Action-CoT), a path-conditioned dual-query training scheme for action generation. Experiments on standard VLN benchmarks show that ReflectVLN improves success rates and path efficiency under a constrained data budget, with favorable training cost and fewer high-level intention calls at inference time, while providing interpretable intermediate decisions for analysis and collaboration. Code is available at: https://github.com/AIprogrammer/ReflectVLN

\end{abstract}    
\section{Introduction}
\label{sec:intro}
Visual-Language Navigation (VLN) in continuous environments is a central problem in embodied AI, in which an agent must follow natural-language instructions to reach a goal through complex real-world scenes. A key challenge lies in long-horizon execution: small control errors accumulate, the agent may drift off route, and successful navigation requires not only short-horizon action execution but also explicit tracking of instruction-level semantic progress and timely plan correction. However, most existing VLN frameworks learn tightly coupled end-to-end policies that map vision and language directly to short-horizon actions, without an explicit mechanism for monitoring sub-goal completion or reflecting on failures during execution.

\begin{wrapfigure}{r}{0.5\textwidth}
  \vspace{-1.1cm}
  \centering
  \includegraphics[width=0.48\textwidth]{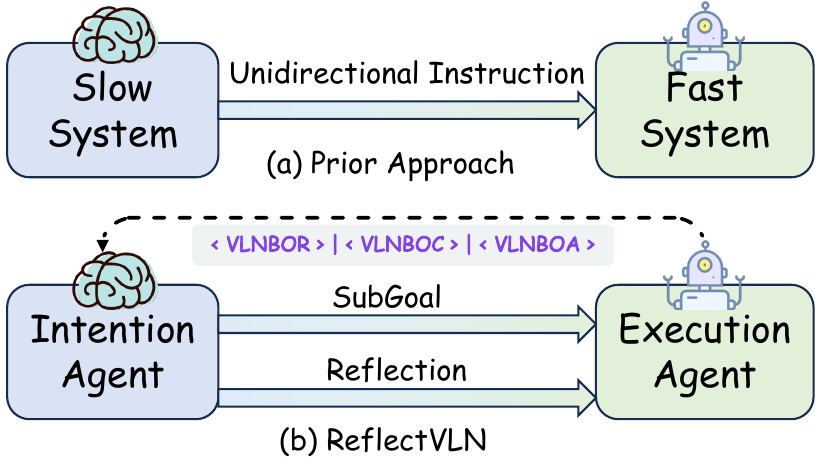}
  % \vspace{-10pt} 
  \caption{\textbf{Comparison between ReflectVLN and prior slow--fast / dual-system VLN frameworks.} Prior cascaded methods pass sub-goals unidirectionally from a slow planner to a fast executor. We define three trigger tokens to explicitly control the system mode. \texttt{<VLNBOR>} denotes the start of the reflection phase for sub-goal generation. \texttt{<VLNBOC>} denotes the start of the correction phase for error diagnosis and recovery. \texttt{<VLNBOA>} denotes the action mode, in which the execution agent acts without invoking the intention agent.}
  \label{fig:contrast}
  \vspace{-10pt} 
\end{wrapfigure}

Prior work often adopts a cascaded design, in which a high-level module predicts sub-goals at a low frequency and a low-level controller executes them. Hi-Robot~\cite{hi_robot} proposes a hierarchical VLA framework with a high-level policy that generates low-level language commands and a low-level policy that outputs actions with verbal responses. Pi-0.5~\cite{pi_0_5} improves generalization via decoupled two-stage training with heterogeneous data sources. DualVLN~\cite{dualNav} further separates high-level reasoning from low-level execution through dual-system training. Despite these advances, existing slow--fast or dual-system designs still rely largely on one-way communication: high-level plans are passed to the fast executor, but execution-time progress and deviation signals are not systematically fed back to the slow system to enable timely sub-goal updates. As a result, the two modules can become semantically misaligned during closed-loop inference, as shown in Fig.~\ref{fig:contrast}(a), leading to delayed replanning and sub-goal switching—an issue that is amplified in long-horizon navigation, where small errors compound. In addition, many paradigms lack an explicit reflection mechanism to detect, diagnose, and revise failures, which limits robustness and recovery in unseen environments.

\textbf{Q1: How can we achieve closed-loop co-evolution between high-level semantic intention and low-level action execution?}
We propose ReflectVLN, as shown in Fig.~\ref{fig:contrast}(b), an agentic VLN framework with bidirectionally interactive intention and execution agents. Compared with existing dual systems, the key distinction lies in the \emph{feedback interface}: the execution agent tracks sub-goal progress and emits deviation signals to trigger the intention agent on demand, rather than relying on one-way or fixed-cadence plan issuance. The intention agent performs deliberate reflection for long-horizon subtask decomposition and correction, while the execution agent conducts short-horizon control conditioned on sub-goals.

\textbf{Q2: How can execution-time error signals be fed back to enable on-demand replanning and reliable recovery?}
To equip the model with reflection capability, we construct a reflection-driven data pipeline. We collect failure cases from model rollouts, invoke an expert policy to generate corrected trajectories, and use a VLM to produce natural-language descriptions of these corrections. An automatic annotation and verification module then filters out hallucinations and inconsistencies, yielding verified error-correction trajectories and reflective feedback for training. Our pipeline enriches corrected trajectories with reflective feedback at the language-level, enabling explicit supervision for recovery-oriented learning. We further introduce Action-CoT, a path-conditioned dual-query training scheme that promotes joint reasoning over a coarse future route and short-horizon actions.

Our contributions can be summarized as follows:
\begin{itemize}
\item We propose ReflectVLN, an agentic VLN framework with bidirectionally interactive intention and execution agents, in which the execution agent not only predicts short-horizon actions but also triggers the intention agent to adaptively update sub-goals via progress and deviation feedback.
\item We propose a two-stage training framework that couples reflective supervision with Action-CoT, encouraging explicit route-level auxiliary reasoning alongside short-horizon control.
\item We develop a reflection-driven data pipeline that converts rollout failures into verified error-correction trajectories with reflective language feedback for recovery-oriented training.
\item We demonstrate competitive improvements on R2R-CE and RxR-CE under a constrained data budget, without large-scale pretraining corpora or multi-round data augmentation, while on-demand intention calls reduce high-level invocations relative to dense periodic refreshing.
\end{itemize}

\section{Related Work}
\label{sec:related-work}

\paragraph{\textbf{Robust Vision-Language Navigation.}}
In recent years, researchers have increasingly focused on the robustness of VLN agents \cite{Envedit, Flame, Human_neural} under distribution shifts \cite{inspiring, Navmorph, Se-vln}. In real-world deployment \cite{BridgeNav, UrbanNav} and long-horizon instruction-following scenarios \cite{dualNav, navforesee}, execution errors can cause an agent to deviate from the expert trajectory, thereby compounding errors and reducing navigation success rates. One prevailing paradigm adopts DAgger \cite{dagger_dataset} and its variants \cite{Envedit, Hg-dagger, Iterative_Correction, Diff-dagger}, which iteratively collect on-policy data and query an expert for ground-truth actions. CorrectNav \cite{Correctnav} further transforms model-generated error trajectories into training data, internalizing error-correction capabilities through a multi-round self-correcting flywheel. However, such methods often rely on high-frequency iterative sampling, which is computationally expensive and difficult to scale. In addition, BudVLN \cite{BudVLN} and DV-VLN \cite{DV-VLN} employ historical re-anchoring and dual verification for rectification, while memory-based methods such as SkillNav \cite{Skill-Nav, Omninav, streamvln} mitigate error propagation through modular atomic skills. Complementary slow--fast or dual-system designs \cite{dualNav} decouple high-level planning from low-level control, yet remain largely one-way: plans are issued to the executor without systematically feeding execution-time progress and failure signals back for timely sub-goal updates. Nevertheless, these approaches still lack a closed-loop recovery mechanism for handling drift during long-horizon execution. In contrast, we propose ReflectVLN, an agentic framework with bidirectionally interactive intention and execution agents for on-demand reflection and deviation correction, trained with a one-round reflective augmentation pipeline under a constrained data budget.

\paragraph{\textbf{Chain-of-Thought (CoT) Reasoning.}}
The paradigm of CoT reasoning, originally designed to enhance the reasoning capabilities of LLMs through explicit step-by-step derivation \cite{cot_for_LLM}, has undergone substantial architectural diversification. While early advances such as self-consistency \cite{Self-consistency} and hierarchical task decomposition \cite{Least-to-most} focused on improving the robustness of internal reasoning, the emergence of VLA models has shifted attention toward grounded decision-making. Strategic exploration in the reasoning space, as enabled by tree-structured search processes \cite{TreeCot, LLMCot}, provides a foundation for more deliberate planning in complex environments. Recent advances have further moved from purely linguistic rationales to multimodal imaginative reasoning. CoT-VLA \cite{Cot-vla} and DreamVLA \cite{Dreamvla} employ visual foresight by predicting future states as an intermediate reasoning modality, while Visual Thoughts \cite{Visual-thoughts} integrates such reasoning into a unified multimodal framework. Moving beyond static prediction, recent methods such as ACoT-VLA \cite{ACoT-VLA} and RoboticCoT \cite{RoboticCoT} prioritize action-centric reasoning chains to ensure temporal consistency in robotic control \cite{Cosmos-reason1}. In contrast, our Action-CoT is a path-conditioned dual-query auxiliary training scheme for VLN short-horizon control, coupled with the intention--execution closed loop rather than used as a standalone sequential language CoT.

\section{Method}
\label{sec:method}
\begin{figure*}[tb]
  \centering
  \includegraphics[width=1.0\linewidth]{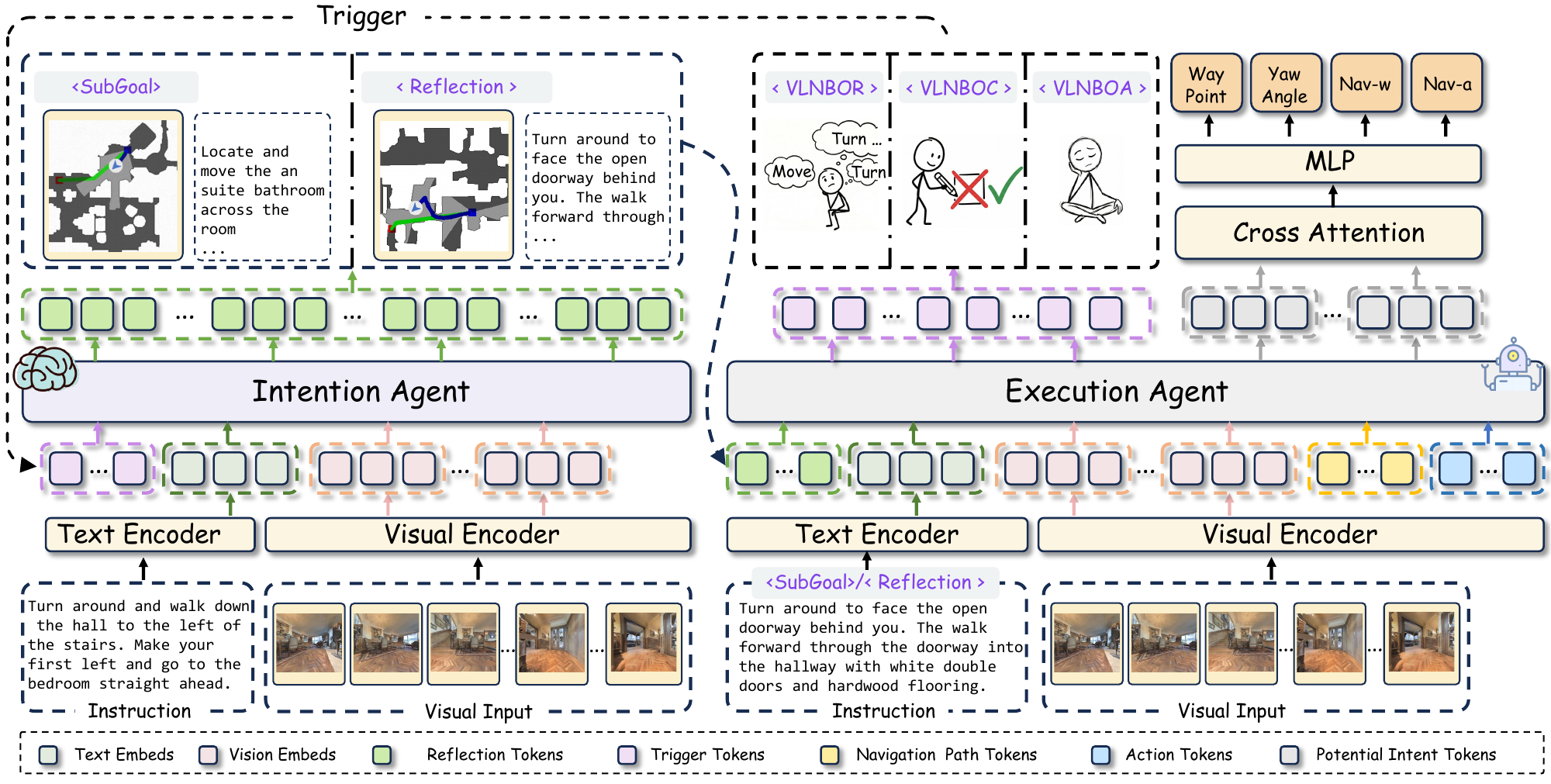}
  \caption{\textbf{Overview of ReflectVLN Framework.} The intention agent operates as a high-level reflective planner that takes egocentric visual history, the natural-language instruction, and execution feedback from the execution agent as input, and produces a reflective sub-goal description summarizing the next navigational intent. Conditioned on this sub-goal together with visual observations, the execution agent outputs continuous motion trajectories and a status token that governs subsequent intention--execution interaction.}
  \label{Fig.framework}
\end{figure*}

We introduce ReflectVLN, an agentic VLN framework with bidirectionally interactive intention and execution agents, as shown in Fig.~\ref{Fig.framework}. The core idea is to couple an execution agent for short-horizon action control with an intention agent for reflective planning and correction.

Concretely, we first establish a formal definition of the VLN task and its inference pipeline (Sec.~\ref{sec:problem_definition}). To provide recovery-oriented supervision, we curate a reflective training set that incorporates off-route scenarios and corresponding recovery trajectories (Sec.~\ref{sec:dataset_construction}). We then detail the intention--execution dual-agent architecture together with Action-CoT for path-conditioned execution (Sec.~\ref{sec:dual_system}), followed by the two-stage training objective (Sec.~\ref{sec:training}) and closed-loop inference procedure (Sec.~\ref{sec:inference_trigger}).

\subsection{Problem Definition}
\label{sec:problem_definition}
VLN agents are generally trained using diverse navigation demonstrations or within simulated environments. Given a natural-language instruction $\ell$, an embodied agent is tasked with navigating to a target goal by strictly following a described route in previously unseen environments.
In our formulation, observations are restricted to monocular visual inputs consisting of current and historical RGB frames $\mathbf{I}_{t-H:t}=\{I_{t-H},\ldots,I_t\}$. Unlike conventional methods, our model operates without depth information, multi-view panoramas, or additional sensors. At each time step $t$, the agent predicts a sequence of relative waypoints that specifies the next motion segment.

The central difficulty is that the instruction $\ell$ describes a global route, whereas the observation history only reveals a local and potentially ambiguous portion of that route. We therefore introduce a language sub-goal $c_t$ as a persistent intermediate state between global instruction understanding and local control. Rather than regenerating $c_t$ at every step, the agent maintains it while execution remains valid and updates it only after detecting sub-goal completion or route deviation. This formulation turns high-level replanning into an event-driven decision and makes the source of each update explicit.

\subsection{Reflective Data Construction}
\label{sec:dataset_construction}
\begin{figure*}[h]
  \centering
  \includegraphics[width=1.0\linewidth]{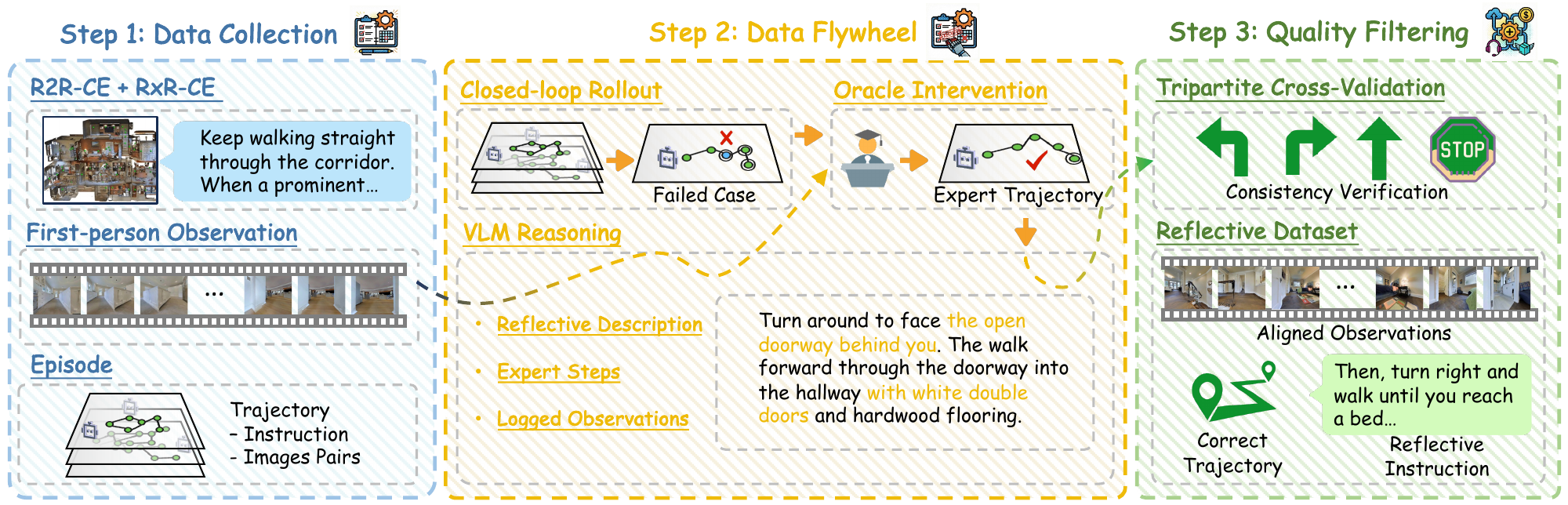}
  \caption{\textbf{Overview of reflective data generation.} The data pipeline contains three steps. (1) Expert/oracle trajectory collection: reference demonstrations and milestone progressions from an oracle expert provide ground-truth trajectories and waypoint-level supervision. (2) Reflective data annotation: we roll out the agent policy, detect deviations from the reference trajectory, and use a VLM to generate a reflective description of sub-goal corrections. (3) Quality filtering: we filter reflective samples by consistency between atomic actions and the reflective annotation.}
  \label{Fig.datapipeline}
\end{figure*}

To mitigate compounding errors in sequential decision-making, we design a reflection-driven data pipeline that transforms navigation failures into targeted training supervision, as shown in Fig.~\ref{Fig.datapipeline}. Importantly, this pipeline is implemented as a \emph{one-round} augmentation procedure rather than a multi-round iterative flywheel: we roll out the current policy once, collect failures, generate corrections, and merge the verified samples into the training set.

\paragraph{\textbf{Expert Data.}}
We build our training set by directly reusing the public expert demonstrations from NavForSee~\cite{navforesee} on R2R-CE~\cite{R2R-CE} and RxR-CE~\cite{RxR-CE}, without collecting new trajectories. Each episode consists of the original instruction paired with the full egocentric RGB observation sequence. The expert data contain approximately 1.3M training samples from RxR-CE and 0.2M from R2R-CE.

\paragraph{\textbf{Reflection-Augmented Recovery Data.}}
During exploration, the current policy is executed in the simulator. When the agent fails to reach the goal or severely leaves the intended path, we invoke a privileged expert with access to the environment graph to compute a recovery trajectory consisting of expert steps, poses, and actions. A VLM (Qwen3-VL-32B~\cite{qwen3technicalreport}) then generates a natural-language trajectory description that explains what should have been done and highlights key visual cues.
To reduce hallucinations, a checker module cross-validates the VLM description against expert steps and logged observations, and discards or repairs inconsistent samples. The checker outputs a structured package with the corrected trajectory, a reflective instruction, and aligned observations. These verified examples are merged into the training set for a single fine-tuning stage on previously failing patterns.
This pipeline yields approximately 100k closed-loop samples. We subsample dominant straight-move transitions and augment stop-containing episodes to balance the supervision distribution.

\subsection{Bidirectionally Interactive Intention--Execution Agents}
\label{sec:dual_system}
ReflectVLN separates semantic deliberation from continuous control while coupling them through an explicit feedback interface. The intention agent translates the global instruction and recent execution history into a language sub-goal, whereas the execution agent grounds this sub-goal into short-horizon controls and emits a discrete status token. This token either maintains action execution or requests a regular or corrective intention update, thereby forming a closed-loop intention--execution process rather than a one-way planning cascade.

The two agents adopt the same Qwen2.5-VL-3B architecture~\cite{Qwen2.5-VL} but are instantiated with independent parameters, denoted by $\theta_{\mathrm{int}}$ and $\theta_{\mathrm{exe}}$. For compactness, we write $\mathrm{R}:=\langle\mathrm{VLNBOR}\rangle$, $\mathrm{C}:=\langle\mathrm{VLNBOC}\rangle$, and $\mathrm{A}:=\langle\mathrm{VLNBOA}\rangle$. Their interaction at time $t$ is summarized as

\begin{equation}
\begin{aligned}
c_t &= \pi_{\theta_{\mathrm{int}}}
    (\ell,\mathbf{I}_{t-H:t},z_t),
    &&z_t\in\{\mathrm{R},\mathrm{C}\},\\
(\hat{P}_t,\hat{A}_t,z_t)
    &= \pi_{\theta_{\mathrm{exe}}}
    (\ell,c_t,\mathbf{I}_{t-H:t}),
\end{aligned}
\label{eq:agent_interface}
\end{equation}
where $c_t$ is a language sub-goal, $\hat{P}_t$ and $\hat{A}_t$ are coarse-route and short-horizon control predictions, and $z_t$ is the execution status.

\subsubsection{Intention Agent}
\label{sec:intention_agent}
The intention agent is a high-level reflective planner. Given the navigation instruction, recent observations, and an update request from the execution agent, it performs one of two language-generation tasks. Upon a regular request ($z_t=\mathrm{R}$), it aligns the global instruction with the observed trajectory and generates the next executable sub-goal. Upon a correction request ($z_t=\mathrm{C}$), it diagnoses off-track behavior and generates a corrective sub-goal. Unlike hierarchical systems that communicate through pixel goals or bounding boxes, our language interface directly exploits the VLM's multimodal reasoning and generation capabilities and requires no additional goal detector.

\subsubsection{Execution Agent with Action-CoT}
\label{sec:execution_agent}
\label{sec:action_cot}
The execution agent maps the current language sub-goal to continuous controls while monitoring whether execution remains consistent with that sub-goal. At each step, it predicts a sequence of $N_q$ controls, each parameterized as
\begin{equation}
\mathbf{u}_t=
\big(\Delta x_t,\Delta y_t,\cos\phi_t,\sin\phi_t,s_t\big),
\label{eq:control}
\end{equation}
where $(\Delta x_t,\Delta y_t)$ is the relative planar displacement, $(\cos\phi_t,\sin\phi_t)$ represents heading, and $s_t$ is a stop logit.

To improve route-level consistency of local control, we equip the execution agent with \emph{Action Chain-of-Thought (Action-CoT)}, a path-conditioned dual-query architecture. Unlike the sequential action-intent formulation of ACoT-VLA~\cite{ACoT-VLA}, ``thought'' in our design is a latent coarse route rather than a token-by-token language rationale. Given the multimodal context $X$ encoded from $(\ell,c_t,\mathbf{I}_{t-H:t})$, navigation queries $Q_{\mathrm{nav}}$ first predict a longer-horizon route:
\begin{equation}
E_{\mathrm{nav}}=f_{\theta_{\mathrm{exe}}}(X,Q_{\mathrm{nav}}),
\qquad
\hat{P}_t=\mathrm{MLP}_{\mathrm{nav}}(E_{\mathrm{nav}}).
\label{eq:qnav}
\end{equation}
The route representation then conditions the short-horizon action queries $Q_{\mathrm{act}}$:
\begin{equation}
E_{\mathrm{act}}=f_{\theta_{\mathrm{exe}}}
\big(X,[E_{\mathrm{nav}};Q_{\mathrm{act}}]\big),
\qquad
\hat{A}_t=\mathrm{MLP}_{\mathrm{act}}(E_{\mathrm{act}}).
\label{eq:qact}
\end{equation}
Both query sets are learnable. The coarse route provides an explicit intermediate prediction that regularizes near-term control toward globally consistent motion. In addition to waypoints, the execution agent autoregressively emits one of three status tokens:
$\langle\mathrm{VLNBOA}\rangle$ continues action execution,
$\langle\mathrm{VLNBOR}\rangle$ indicates completion of the current sub-goal and requests the next regular sub-goal, and
$\langle\mathrm{VLNBOC}\rangle$ indicates deviation and requests corrective reflection. These outputs provide the feedback interface from execution to intention.

\subsection{Two-Stage Training}
\label{sec:training}
We train the independently parameterized agents with complementary objectives. Stage I initializes the execution agent through Action-CoT pretraining on expert trajectories. Stage II trains the intention agent on regular and corrective language supervision while continuing to optimize the execution agent for waypoint and status-token prediction. No gradient is propagated between the two agents.

\subsubsection{Stage I: Action-CoT Pretraining}
Both the short-horizon action sequence $\hat{A}$ and the coarse route $\hat{P}$ use the parameterization in Eq.~\eqref{eq:control}. For a predicted sequence $\hat{U}=\{\hat{\mathbf{u}}_j\}_{j=1}^{N}$ and its expert target $U^*$, we define
\begin{align}
\mathcal{L}_{\mathrm{pos}}(\hat U,U^*)
&=\frac{1}{N}\sum_{j=1}^{N}
\left(\lvert\Delta x_j-\Delta x_j^*\rvert+
\lvert\Delta y_j-\Delta y_j^*\rvert\right), \label{eq:lpos}\\
\mathcal{L}_{\mathrm{ang}}(\hat U,U^*)
&=1-\frac{1}{N}\sum_{j=1}^{N}
\left(\cos\phi_j\cos\phi_j^*+
\sin\phi_j\sin\phi_j^*\right), \label{eq:lang}\\
\mathcal{L}_{\mathrm{stop}}(\hat U,U^*)
&=-\frac{1}{N}\sum_{j=1}^{N}
\left[s_j^*\log\sigma(s_j)
+(1-s_j^*)\log(1-\sigma(s_j))\right].
\label{eq:lstop}
\end{align}
Their weighted combination is
\begin{equation}
\mathcal{L}_{\mathrm{traj}}(\hat U,U^*)
=\lambda_{\mathrm{pos}}\mathcal{L}_{\mathrm{pos}}
+\lambda_{\mathrm{ang}}\mathcal{L}_{\mathrm{ang}}
+\lambda_{\mathrm{stop}}\mathcal{L}_{\mathrm{stop}}.
\label{eq:ltraj}
\end{equation}
The short-horizon target $A^*$ contains the next $N_q$ expert controls. The coarse-route target $P^*$ is obtained by uniformly sampling future expert waypoints over a longer horizon. Action-CoT pretraining minimizes
\begin{equation}
\mathcal{L}_{\mathrm{WP}}
=\mathcal{L}_{\mathrm{traj}}(\hat A,A^*)
+\beta\mathcal{L}_{\mathrm{traj}}(\hat P,P^*),
\label{eq:ltotal}
\end{equation}
where the second term encourages navigation queries to capture route structure before producing local controls. We use $\lambda_{\mathrm{pos}}=\lambda_{\mathrm{ang}}=\lambda_{\mathrm{stop}}=1$ and $\beta=0.1$.

\subsubsection{Stage II: Status-Aware Fine-Tuning}
\paragraph{Intention objective.}
For an intention request $z\in\{\mathrm{R},\mathrm{C}\}$, the intention agent is trained by autoregressive imitation:
\begin{equation}
\mathcal{L}_{\mathrm{int}}^{(z)}
=-\sum_{j=1}^{T}
\log p_{\theta_{\mathrm{int}}}
\left(y_j^*\mid y_{<j}^*,\ell,\mathbf{I}_{t-H:t},z\right),
\label{eq:lint}
\end{equation}
where $y^*$ is the target regular or corrective sub-goal. This objective trains semantic sub-goal generation only and is distinct from the continuous action objective.

\paragraph{Execution objective.}
The execution agent learns the status output with an autoregressive classification loss
\begin{equation}
\mathcal{L}_{\mathrm{status}}
=-\log p_{\theta_{\mathrm{exe}}}
\left(z_t^*\mid\ell,c_t,\mathbf{I}_{t-H:t}\right).
\label{eq:lstatus}
\end{equation}
Together with Action-CoT waypoint supervision, its objective is
\begin{equation}
\mathcal{L}_{\mathrm{exe}}
=\mathbf{1}[z_t^*=\mathrm{A}]\,\mathcal{L}_{\mathrm{WP}}
+\lambda_{\mathrm{status}}\mathcal{L}_{\mathrm{status}},
\label{eq:lexe}
\end{equation}
where waypoint loss is applied when the target status continues execution. The complete objective is
\begin{equation}
\min_{\theta_{\mathrm{int}},\theta_{\mathrm{exe}}}
\quad
\sum_{z\in\{\mathrm{R},\mathrm{C}\}}
\rho_z\,\mathbb{E}_{\xi\sim\mathcal{D}_z}
\left[\mathcal{L}_{\mathrm{int}}^{(z)}\right]
+\rho_{\mathrm{exe}}\,
\mathbb{E}_{\xi\sim\mathcal{D}_{\mathrm{exe}}}
\left[\mathcal{L}_{\mathrm{exe}}\right],
\label{eq:full_objective}
\end{equation}
where $\rho$ denotes the sampling weight of each data source.

\subsection{Closed-Loop Inference}
\label{sec:inference_trigger}
At test time, the intention agent first generates an initial sub-goal. Thereafter, the status token emitted by the execution agent directly determines whether to execute the predicted control or invoke the intention agent for an update. The resulting event-driven interaction requires no oracle trigger and limits expensive intention calls while enabling timely recovery; fixed-interval alternatives are compared in Sec.~\ref{sec:ablation}.

\begin{algorithm}[t]
\caption{Closed-loop intention--execution inference}
\label{alg:inference}
\begin{algorithmic}[1]
\Require instruction $\ell$
\State $c\gets\textsc{IntentionAgent}(\ell,\varnothing,\mathrm{R})$
\While{not stopped}
    \State $(\hat P_t,\hat A_t,z_t)\gets
    \textsc{ExecutionAgent}(\ell,c,\mathbf{I}_{t-H:t})$
    \If{$z_t=\mathrm{C}$}
        \State $c\gets\textsc{IntentionAgent}
        (\ell,\mathbf{I}_{t-H:t},\mathrm{C})$
    \ElsIf{$z_t=\mathrm{R}$}
        \State $c\gets\textsc{IntentionAgent}
        (\ell,\mathbf{I}_{t-H:t},\mathrm{R})$
    \Else
        \State execute the first control in $\hat A_t$
        \If{stop is predicted}
            \State \textbf{break}
        \EndIf
    \EndIf
\EndWhile
\end{algorithmic}
\end{algorithm}

\paragraph{\textbf{Implementation details.}}
Both agents are initialized from separate Qwen2.5-VL-3B checkpoints. The intention agent retains the language-generation head, whereas the execution agent adds the navigation-query and action-query MLP heads and generates the status token through its language head. Observations use a history of $H$ frames, and the action branch predicts $N_q$ future controls at each step. 

\section{Experiments}
\label{sec:experiments}

\subsection{Experimental Setup}
\label{sec:exp_setup}
Both agents are initialized from Qwen2.5-VL-3B and trained following the two-stage procedure in Fig.~\ref{Fig.framework}. In Stage~1, we pretrain the execution agent with Action-CoT on expert trajectories for one epoch. In Stage~2, we train the decoupled agents for one epoch using a mixture of expert demonstrations and one-round reflective trajectories. We use a learning rate of $1\times10^{-5}$ in both stages. Training is conducted on 32 NVIDIA H20 GPUs and takes approximately 16 and 60 hours for the two stages, respectively.

\paragraph{\textbf{Benchmarks and metrics.}}
We evaluate on the Val-Unseen splits of R2R-CE and RxR-CE, two standard continuous-control VLN benchmarks built on Matterport3D. We report navigation error (NE), success rate (SR), oracle success rate (OSR), and success weighted by path length (SPL). An episode is successful if the agent stops within 3\,m of the goal; OSR instead measures whether the trajectory visits this region at any point.

\subsection{Comparison with State of the Art}
\label{sec:comparison}

\begin{table*}[t]
\centering
\caption{\textbf{Comparison with existing approaches on the VLN-CE benchmarks (R2R and RxR Val-Unseen split).} Pano, Odo, Depth, and S.RGB respectively represent panoramic view, odometry, depth, and single RGB. \textbf{CorrectNav$^{*}$} denotes the result obtained using only the first round of self-correction flywheel post-training.}
\label{tab:r2r_rxr_valunseen}
\begin{adjustbox}{width=\textwidth}
\small
\setlength{\tabcolsep}{4pt}
\renewcommand{\arraystretch}{1.05}
\begin{tabular}{lccccccccccc}
\toprule
\multirow{2}{*}{Method} &
\multicolumn{4}{c}{Observation} &
\multicolumn{4}{c}{R2R-CE Val-Unseen} &
\multicolumn{3}{c}{RxR-CE Val-Unseen} \\
\cmidrule(lr){2-5}\cmidrule(lr){6-9}\cmidrule(lr){10-12}
& S.RGB & Pano. & Depth & Odo &
NE$\downarrow$ & OS$\uparrow$ & SR$\uparrow$ & SPL$\uparrow$ &
NE$\downarrow$ & SR$\uparrow$ & SPL$\uparrow$ \\
\midrule
HPN+DN~\cite{hpn_dn}         &      & \checkmark & \checkmark & \checkmark & 6.31 & 40.0 & 36.0 & 34.0 & --   & --   & --   \\
CMA~\cite{cma}               &      & \checkmark & \checkmark & \checkmark & 6.20 & 52.0 & 41.0 & 36.0 & 8.76 & 26.5 & 22.1 \\
Sim2Sim~\cite{sim2sim}       &      & \checkmark & \checkmark & \checkmark & 6.07 & 52.0 & 43.0 & 36.0 & 8.76 & 26.5 & 22.1 \\
GridMM~\cite{gridmm}         &      & \checkmark & \checkmark & \checkmark & 5.11 & 61.0 & 49.0 & 41.0 & --   & --   & --   \\
DreamWalker~\cite{dreamwalker}&     & \checkmark & \checkmark & \checkmark & 5.53 & 59.0 & 49.0 & 44.0 & --   & --   & --   \\
Reborn~\cite{reborn}         &      & \checkmark & \checkmark & \checkmark & 5.40 & 57.0 & 50.0 & 46.0 & 5.98 & 48.6 & 42.0 \\
ETPNav~\cite{etpnav}         &      & \checkmark & \checkmark & \checkmark & 4.71 & 65.0 & 57.0 & 49.0 & 5.64 & 54.7 & 44.8 \\
HNR~\cite{hnr}               &      & \checkmark & \checkmark & \checkmark & 4.42 & 67.0 & 61.0 & 51.0 & 5.50 & 56.3 & 46.7 \\
\midrule
AG-CMTP~\cite{cmtp}          &      & \checkmark & \checkmark & \checkmark & 7.90 & 39.0 & 23.0 & 19.0 & --   & --   & --   \\
R2R-CMTP~\cite{cmtp}         &      & \checkmark & \checkmark & \checkmark & 7.90 & 38.0 & 26.0 & 22.0 & --   & --   & --   \\
Instruct-Nav~\cite{instructnav} &   & \checkmark & \checkmark & \checkmark & 6.89 & --   & 31.0 & 24.0 & --   & --   & --   \\
LAW~\cite{law}               & \checkmark &      & \checkmark & \checkmark & 6.83 & 44.0 & 35.0 & 31.0 & 10.90& 8.0  & 8.0  \\
CM2~\cite{cm2}               & \checkmark &      & \checkmark & \checkmark & 7.02 & 41.0 & 34.0 & 27.0 & --   & --   & --   \\
WS-MGMap~\cite{wsmgmap}      & \checkmark &      & \checkmark & \checkmark & 6.28 & 47.0 & 38.0 & 34.0 & --   & --   & --   \\
AO-Planner~\cite{aoplanner}  &      & \checkmark & \checkmark &          & 5.55 & 59.0 & 47.0 & 33.0 & --   & --   & --   \\
Seq2Seq~\cite{seq2seq}       & \checkmark &      & \checkmark &          & 7.77 & 37.0 & 25.0 & 22.0 & 12.10& 13.9 & 11.9 \\
CMA~\cite{seq2seq}           & \checkmark &      & \checkmark &          & 7.37 & 40.0 & 32.0 & 30.0 & --   & --   & --   \\
NAVid~\cite{navid}           & \checkmark &      &          &          & 5.47 & 49.0 & 37.0 & 35.0 & --   & --   & --   \\
UniNAVid~\cite{uninavid}     & \checkmark &      &          &          & 5.58 & 53.5 & 47.0 & 42.7 & 6.24 & 48.7 & 40.9 \\
NaViLA~\cite{Navila}         & \checkmark &      &          &          & 5.22 & 62.5 & 54.0 & 49.0 & 6.77 & 49.3 & 44.0 \\
StreamVLN~\cite{streamvln}   & \checkmark &      &          &          & 4.98 & 64.2 & 56.9 & 51.9 & 6.22 & 52.9 & 46.0 \\
CorrectNav$^{*}$~\cite{Correctnav}             & \checkmark &      &          &          & 4.50 & --   & 61.4 & \textbf{59.0} & 4.40 & 63.1 & 57.0 \\
% CorrectNav~\cite{Correctnav} & \checkmark &      &          &          & 4.24 & 67.5 & \textbf{65.1} & \textbf{62.3} & 4.09 & \textbf{69.3} & \textbf{63.3} \\
DualVLN~\cite{dualNav}       & \checkmark &      &          &          & \textbf{4.05} & \textbf{70.7} & \textbf{64.3} & 58.5 & 4.58 & 61.4 & 51.8 \\
\midrule
ReflectVLN (Ours)            & \checkmark &      &          &          & 4.19 & 67.3 & 62.8 & 58.5 & \textbf{3.98} & \textbf{66.0} & \textbf{57.2} \\
\bottomrule
\end{tabular}
\end{adjustbox}
\end{table*}

\begin{figure}[tb]
  \centering
  \includegraphics[width=\linewidth,height=0.22\textheight]{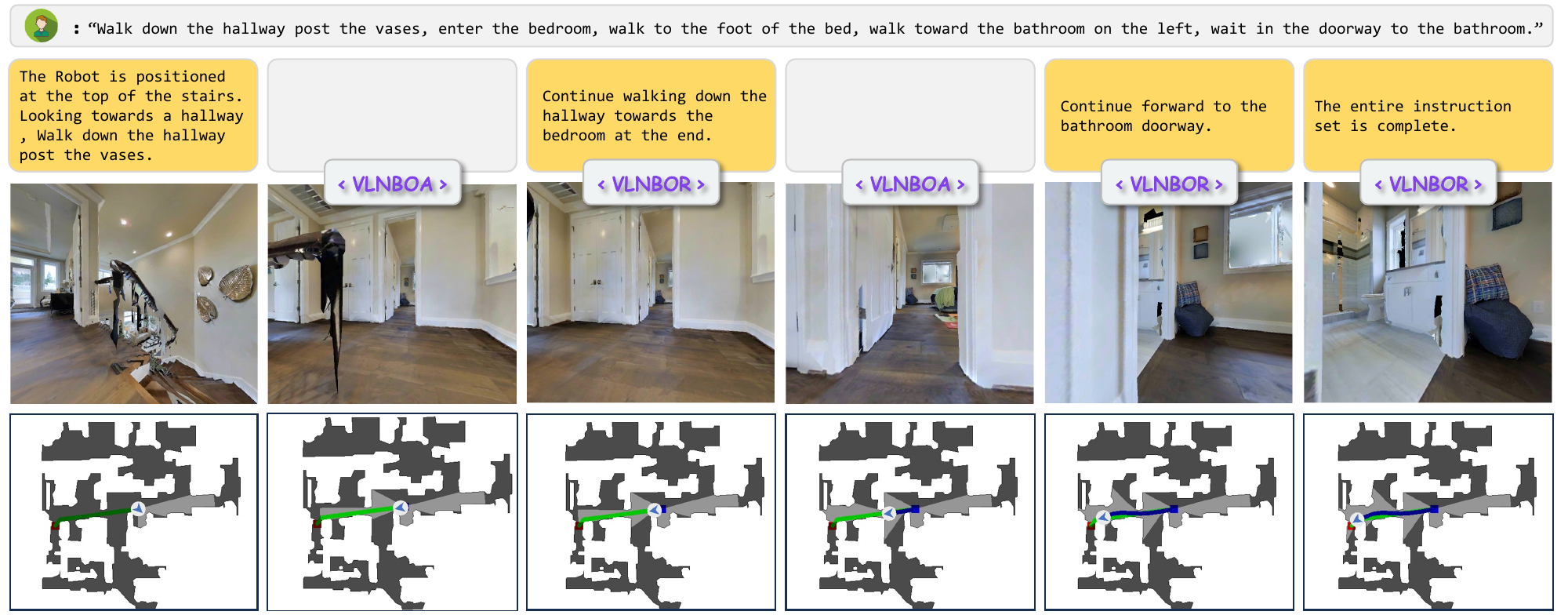}\par\smallskip
  \includegraphics[width=\linewidth,height=0.22\textheight]{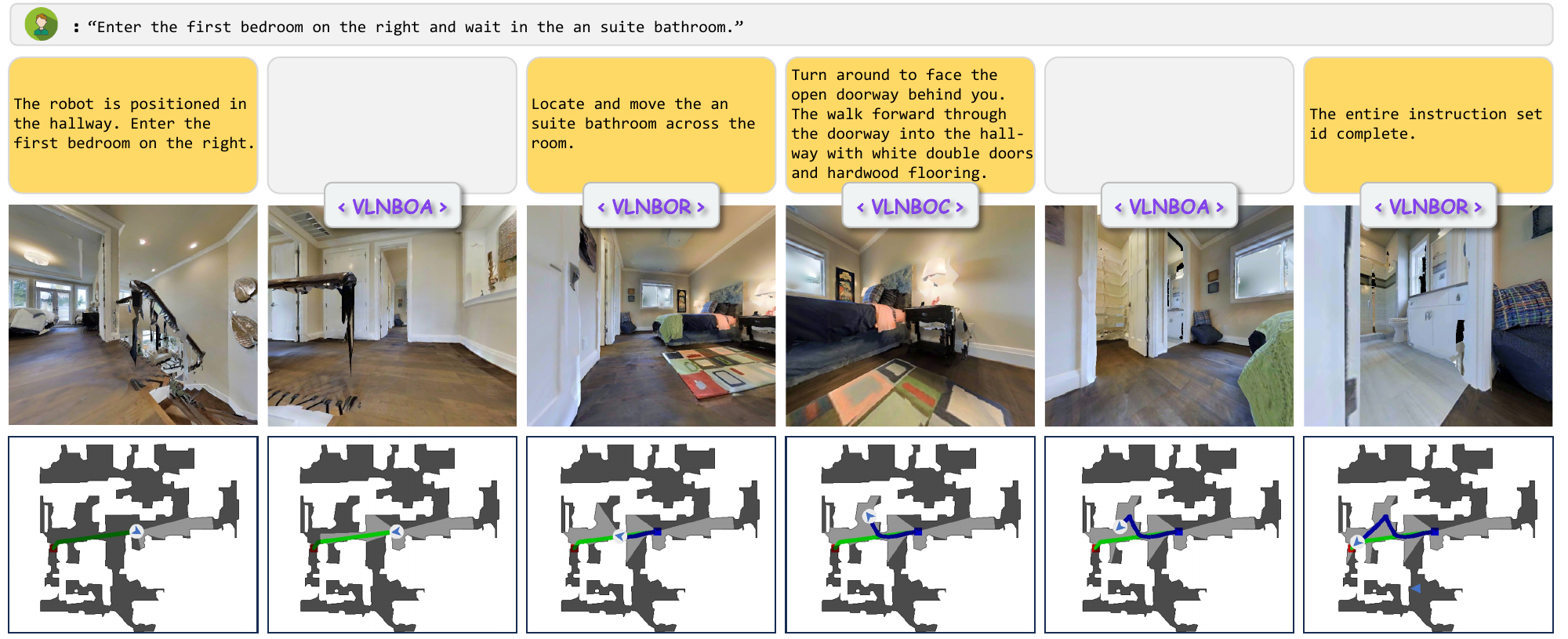}
  \caption{\textbf{Example visualizations on R2R Val-Unseen.} Top: sub-goal grounding and progress tracking. Bottom: off-track detection and recovery via closed-loop replanning.}
  \label{fig:exp}
\end{figure}

\paragraph{\textbf{Main results.}}
Table~\ref{tab:r2r_rxr_valunseen} compares ReflectVLN with existing VLN-CE approaches. ReflectVLN achieves strong performance using only monocular RGB observations, without panoramic views, depth, or odometry. Compared with the RGB-only StreamVLN baseline, it improves SR by 5.9 percentage points on R2R and 13.1 points on RxR, with corresponding gains in SPL. The larger improvement on RxR, which contains longer and more detailed instructions, supports the benefit of maintaining semantic progress and revising sub-goals during long-horizon execution.

ReflectVLN is particularly effective on RxR: it obtains the lowest NE among the compared methods and outperforms DualVLN by 4.6 SR points and 5.4 SPL points. On R2R, it matches DualVLN in SPL while remaining slightly lower in SR. Under the one-round setting of CorrectNav, ReflectVLN yields better SR on both benchmarks and comparable or better SPL. These results show that closed-loop intention--execution interaction provides a favorable accuracy--data trade-off, rather than uniformly dominating methods trained with larger-scale or iterative supervision.

\paragraph{\textbf{Training efficiency.}}
\begin{table}[t]
\centering
\small
\setlength{\tabcolsep}{4pt}
\begin{tabular}{l l c}
\toprule
Method & Data sources & Samples \\
\midrule
CorrectNav (7B) & R2R+RxR+ScaleVLN+VQA & $\sim$3M (1st round) \\
DualVLN (7B) & R2R+RxR+ScaleVLN & $\sim$12M \\
Ours (2$\times$3B) & R2R+RxR+reflection & $\sim$1.6M \\
\bottomrule
\end{tabular}
\caption{Training data scale. ReflectVLN uses two independently parameterized 3B agents and approximately 1.5M expert plus 100k reflection-augmented samples, without ScaleVLN.}
\label{tab:efficiency}
\end{table}
As summarized in Table~\ref{tab:efficiency}, ReflectVLN uses a 3B backbone and 1.5M samples drawn only from R2R and RxR. In contrast, DualVLN and CorrectNav use 7B backbones, additional ScaleVLN data, and substantially more training samples or iterative post-training. The competitive results under this controlled budget indicate that the gains arise from more effective reflective supervision and agent interaction, rather than data or model scaling alone.

\paragraph{\textbf{Qualitative analysis.}}
Figure~\ref{fig:exp} shows representative R2R Val-Unseen rollouts. The intention agent emits sub-goal descriptions that decompose long-horizon instructions into actionable steps, making semantic progress inspectable. When the execution agent drifts, progress/deviation signals trigger on-demand reflection, which revises the sub-goal and enables recovery to the correct route instead of continuing down an incorrect branch.

\subsection{Ablation Studies}
\label{sec:ablation}
\begin{table*}[htbp]
\centering
\caption{Ablation of decoupled vs.\ unified training on R2R-CE Val-Unseen.}
\label{tab:ablation_onetwovla}
\begin{adjustbox}{width=0.5\textwidth}
\setlength{\tabcolsep}{6pt}
\renewcommand{\arraystretch}{1.15}
\begin{tabular}{l|cccc}
\toprule
\multirow{2}{*}{methods} &
\multicolumn{4}{c}{R2R Validation Unseen} \\
\cmidrule(lr){2-5}
& NE$\downarrow$ & OS$\uparrow$ & SR$\uparrow$ & SPL$\uparrow$ \\
\midrule
Unified & 5.052 & 68.6 & 57.3 & 52.7 \\
Ours & 4.190 & 67.3 & 62.8 & 58.5 \\
\bottomrule
\end{tabular}
\end{adjustbox}
\end{table*}

\begin{table}[htbp]
\centering
\caption{Ablation of Action-CoT pretraining on R2R-CE Val-Unseen.}
\label{tab:ablation_acot}
\begin{adjustbox}{width=0.5\textwidth}
\setlength{\tabcolsep}{4.5pt}
\renewcommand{\arraystretch}{1.1}
\resizebox{\columnwidth}{!}{
\begin{tabular}{l|cccc}
\toprule
\multirow{2}{*}{Methods} &
\multicolumn{4}{c}{R2R Validation Unseen} \\
\cmidrule{2-5}
& NE$\downarrow$ & OS$\uparrow$ & SR$\uparrow$ & SPL$\uparrow$ \\
\midrule
Vanilla & 5.145 & 59.5 & 52.8 & 50.1 \\
Action-CoT & 4.727 & 64.9 & 60.0 & 57.2 \\
\bottomrule
\end{tabular}
}
\end{adjustbox}
\end{table}

\begin{table}[htbp]
\centering
\caption{On-demand intention triggering vs.\ fixed intervals on R2R-CE Val-Unseen. Interval denotes steps between intention calls.}
\label{tab:interval}
\setlength{\tabcolsep}{6pt}
\begin{tabular}{l|ccc}
\toprule
Trigger & NE$\downarrow$ & SR$\uparrow$ & SPL$\uparrow$ \\
\midrule
Fixed (every 4) & 4.25 & 61.8 & 57.6 \\
Fixed (every 12) & 4.50 & 59.4 & 56.5 \\
On-demand (Ours) & \textbf{4.19} & \textbf{62.8} & \textbf{58.5} \\
\bottomrule
\end{tabular}
\end{table}

\begin{table}[htbp]
\centering
\caption{Inference chunk settings vs.\ DualVLN on R2R-CE Val-Unseen.}
\label{tab:ablation_model_efficiency}
\begin{adjustbox}{width=0.6\textwidth}
\setlength{\tabcolsep}{10pt}
\renewcommand{\arraystretch}{1.2}
\begin{tabular}{c c c cc}
\toprule
\multirow{2}{*}{methods} & \multirow{2}{*}{Global-Chunk} & \multirow{2}{*}{Local-Chunk} & \multicolumn{2}{c}{R2R Validation Unseen} \\
\cmidrule(lr){4-5}
& & & SR$\uparrow$ & OSR$\uparrow$ \\
\midrule
\multirow{3}{*}{DualVLN} 
& 2  & 4 & 64.9 & 70.9 \\
& 3 & 4 & 64.4 & 71.2 \\
& 2 & 6 & 62.6 & 69.9 \\
\midrule
Ours & 1 & 9.63 & 62.8 & 67.3 \\
\bottomrule
\end{tabular}
\end{adjustbox}
\end{table}

\paragraph{\textbf{Effect of Action-CoT.}}
Table~\ref{tab:ablation_acot} shows that Action-CoT improves SR and SPL by 7.2 and 7.1 points, respectively, while also reducing NE. This consistent improvement across success, efficiency, and goal distance suggests that coarse-route prediction provides useful path-level supervision for short-horizon control.

\paragraph{\textbf{Decoupled agent design.}}
As shown in Table~\ref{tab:ablation_onetwovla}, decoupling intention reasoning from action execution improves SR and SPL by 5.5 and 5.8 points, respectively, over a unified VLM baseline(Qwen2.5-VL-7B). Although the unified model achieves a marginally higher OSR, it is less effective at translating near-goal visits into successful terminations. Notably, this comparison involves distinct parameter budgets (a dual 3B-agent configuration versus a single 7B unified model). The superior performance of our approach, despite a smaller total parameter count, demonstrates that the gains are driven by the complete decoupled design itself, rather than mere architectural separation or parameter scaling.

\paragraph{\textbf{On-demand intention updates.}}
Table~\ref{tab:interval} compares feedback-triggered updates with periodic replanning. Fixed intervals expose a clear trade-off: infrequent updates delay correction, whereas frequent updates invoke the high-level agent unnecessarily. On-demand triggering achieves the best NE, SR, and SPL among the tested schedules, showing that execution feedback provides a more effective update criterion than a fixed temporal cadence.

\subsection{Inference Efficiency}
\label{sec:model_size}
Table~\ref{tab:ablation_model_efficiency} compares high-level invocation frequency with DualVLN~\cite{dualNav}. ReflectVLN invokes its intention agent once every 9.63 execution steps on average, substantially less often than the fixed schedules used by DualVLN, while retaining comparable SR. This result demonstrates reduced reliance on high-level replanning. Because the table measures invocation frequency rather than latency or FLOPs, it should be interpreted as interaction efficiency rather than end-to-end computational efficiency.

\section{Conclusion}
\label{sec:conclusion}
We present ReflectVLN, an agentic VLN framework with bidirectionally interactive intention and execution agents that targets long-horizon error accumulation. The intention agent performs subtask decomposition and corrective planning, while the execution agent conducts short-horizon control with progress and deviation monitoring; closed-loop feedback enables on-demand reflection and recovery. Action-CoT provides path-conditioned auxiliary supervision for temporally coherent action prediction with interpretable intermediate signals. Experiments on standard VLN benchmarks show competitive success and path efficiency under a constrained data and training budget, together with fewer high-level intention calls and interpretable sub-goal decisions that facilitate analysis and collaboration.
The invocation-frequency analysis characterizes interaction efficiency rather than complete system latency or computational cost. In addition, the decoupled and unified configurations differ in total parameter count, so their comparison evaluates the complete system designs rather than architectural separation in isolation. Future work can study capacity-matched agent designs and more direct measures of recovery, triggering quality, and end-to-end efficiency.

% \section*{Acknowledgements}
% Please insert your acknowledgments here.

\bibliographystyle{splncs04}
\bibliography{main}
\end{document}